\documentclass[a4paper]{article}

\usepackage{INTERSPEECH2022}
\usepackage{microtype}
\usepackage{graphicx}
\usepackage{pifont}
\usepackage{multirow}
\usepackage{amsmath}
\usepackage{hyperref}

\title{DUAL: Discrete Spoken Unit Adaptive Learning for Textless Spoken Question Answering}

\name{Guan-Ting Lin$^1$, Yung-Sung Chuang$^2$, Ho-Lam Chung$^1$, Shu-wen Yang$^1$, Hsuan-Jui Chen$^1$, Shuyan Dong$^3$, Shang-Wen Li$^3$, Abdelrahman Mohamed$^3$, Hung-yi Lee$^1$, Lin-shan Lee$^1$}
\address{
  $^1$National Taiwan University, Taiwan\\
  $^2$Massachusetts Institute of Technology, USA\\
  $^3$Meta AI, USA
}
\email{r10942104@ntu.edu.tw, shangwel@fb.com, abdo@fb.com, hungyilee@ntu.edu.tw}

\begin{document}
\maketitle
\begin{abstract}
Spoken Question Answering (SQA) is to find the answer from a spoken document given a question, which is crucial for personal assistants when replying to the queries from the users. Existing SQA methods all rely on Automatic Speech Recognition (ASR) transcripts. Not only does ASR need to be trained with massive annotated data that are time and cost-prohibitive to collect for low-resourced languages, but more importantly, very often the answers to the questions include name entities or out-of-vocabulary words that cannot be recognized correctly. Also, ASR aims to minimize recognition errors equally over all words, including many function words irrelevant to the SQA task. Therefore, SQA without ASR transcripts (textless) is always highly desired, although known to be very difficult.

This work proposes Discrete Spoken Unit Adaptive Learning (DUAL), leveraging unlabeled data for pre-training and fine-tuned by the SQA downstream task. The time intervals of spoken answers can be directly predicted from spoken documents. We also release a new SQA benchmark corpus, NMSQA, for data with more realistic scenarios. We empirically showed that DUAL yields results comparable to those obtained by cascading ASR and text QA model and robust to real-world data. Our code and model will be open-sourced\footnote{\url{https://github.com/DanielLin94144/DUAL-textless-SQA}}.
\end{abstract}
\section{Introduction}
Spoken Question Answering (SQA) aims to find the answer from a spoken document given a question in either text or spoken form. SQA is crucial for personal assistants when replying to the questions from the user's spoken queries. Unlike many spoken language understanding tasks such as speech translation or intent classification, in which the required understanding of semantics is primarily on utterance level, the SQA task requires sophisticated comprehension and reasoning over much longer audio content. In addition to understanding the question and comprehending the global information in the audio context, it also needs to catch the fine-grained information to precisely locate the answer span out of the long audio context. Thus, SQA is known to be a very challenging task.
\begin{figure}[t]
    \centering
    \includegraphics[width=0.45\textwidth]{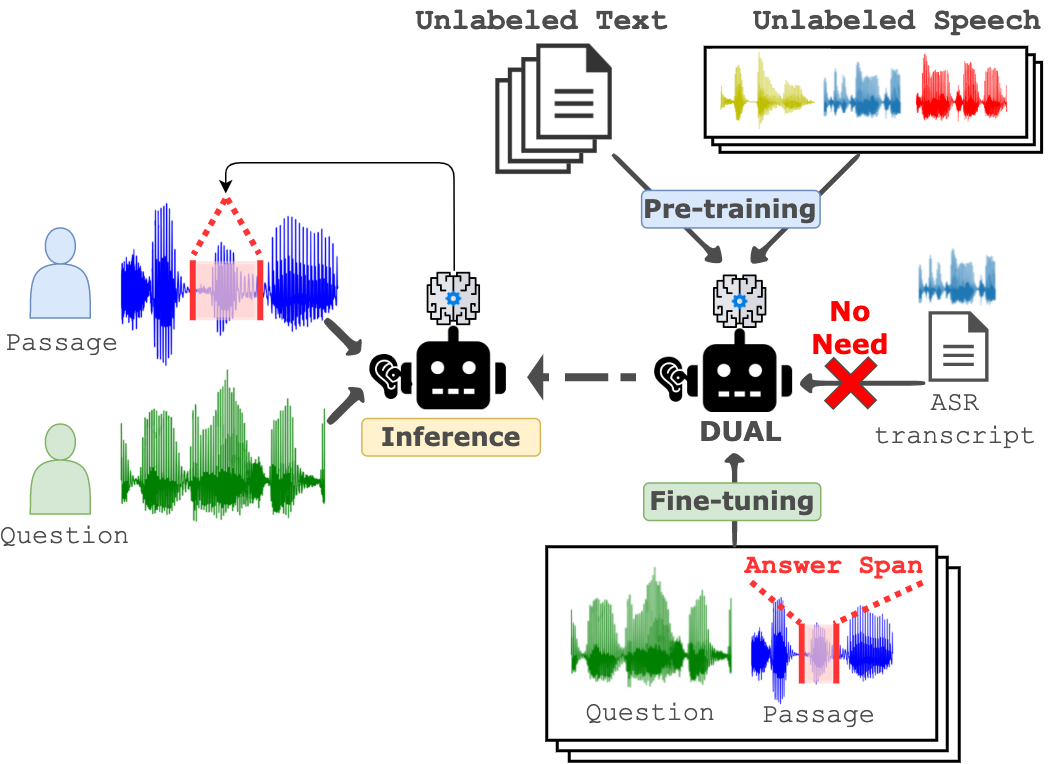}
    \vspace{-0.15cm}
    \caption{The proposed DUAL framework for textless (ASR transcript-free) SQA. All the passages, questions, and answers are in spoken form. The time intervals of the spoken answers can be extracted from the spoken passage without ASR transcripts.}
    \vspace{-0.4cm}
    \label{fig:scenario}
\end{figure}

The conventional SQA system consists of the cascade of an Automatic Speech Recognition (ASR) engine followed by a text-trained QA (TQA) model. However, speech recognition errors naturally cause catastrophic problems to the TQA. \cite{li2018spoken}, and several works \cite{lee2019mitigating, you2021knowledge, You2021MRDNetMR, su2020improving} intended to alleviate such problems by knowledge distillation, including adapting the TQA model to be more robust against recognition errors. Some other efforts \cite{chuang2019speechbert, chung2020splat} exploited paired speech and transcripts to construct a cross-modal speech and text pre-trained model with aligned semantics fine-tuned end-to-end, in which the speech recognition error problems can be mitigated to some degree, and SQA performance improved. 

However, ASR errors remain to be a major problem for SQA tasks. The correct answers to the questions often include name entities or out-of-vocabulary (OOV) words that can never be recognized. The key information is thus inevitably lost when the audio signals are transformed into transcripts with errors, and there is no way to recover them in the following TQA stage. Also, the ASR engine was trained by minimizing the word error rate (WER), which was evaluated equally overall words including many function words irrelevant to the SQA task. So the cascade of two stages (ASR and TQA) individually optimized with two different criteria cannot perform as well as a single-stage global performance goal. So it is highly desired to capture the information directly, rather than from the ASR transcripts, and obtain overall performance not constrained by ASR accuracies. 

Also, ASR engines have to be trained with vast quantities of human-annotated audio-transcript data, which are time-consuming and expensive to collect for the thousands of low-resourced languages over the world, when low and robust enough error rates are considered. Furthermore, there exist many languages without written form worldwide. All the above imply technologies for ASR transcript-free (textless) SQA are highly desired although challenging.

In this work, we propose the first known textless (i.e., ASR transcript-free) SQA framework as in Figure \ref{fig:scenario}. Inspired by the concept of \textit{Textless NLP} \cite{lakhotia2021generative, polyak2021speech, kharitonov2021text, kreuk2021textless, lee2021direct, lee2021textless}, which encodes speech signals into discrete units for modeling, and the \textit{pre-trained language models transferability} of \cite{jurafsky2020learning, kao2021bert, chiang2021transferability, lu2021pretrained, ri2022pretraining}, we propose Discrete Unit Adaptive Learning (DUAL) for textless SQA. DUAL leverages pre-trained models to obtain quantized, length-condensed speech representations from audio signals and further adapts the pre-trained language model to achieve competitive SQA results without any ASR transcripts. The time span of the answer can be directly located from the audio context and played to the user, so the extracted answers do not suffer from speech recognition errors or out-of-vocabulary (OOV) problems because \textit{NO} ASR is performed.

Furthermore, despite the increasing efforts to build SQA benchmark corpora \cite{li2018spoken, tseng2016towards, rajpurkar-etal-2016-squad, you2020towards, lee2018odsqa, ravichander2021noiseqa, faisal2021sd}, there is still a lack of natural and large-scale SQA datasets featuring real-world scenarios. For this purpose, we release a novel benchmark corpus, Natural Multi-speaker Spoken Question Answering (NMSQA). In this corpus, the test set was produced by human speakers, and the training and validation set were synthesized from Amazon Polly TTS service with industrial-grade quality. We also assign two different speakers to read the pairs of passage and question, examining whether our textless SQA system is speaker-independent.

The contributions of this paper are summarized below: 
\begin{itemize}
    \item We propose DUAL as the first known framework for textless SQA, not utilizing ASR transcripts and not suffering from ASR errors. 
    \item We open-source the NMSQA dataset for SQA in real-world scenarios.
    \item DUAL achieved performance competitive to those obtained by cascading ASR and TQA, and significantly better when the word error rate exceeded 30 \%.
    \item DUAL is more robust and retains the performance for the real-speaker testing set, which was not easily achievable for the cascade approach.
\end{itemize}
\section{Method}
\subsection{Problem Formulation}
The form of SQA dataset $D$ is $\{\mathbf{q, p}, a\}$, corresponding to the question $\mathbf{q}$, passage $\mathbf{p}$, and answer $a$, all in spoken form in this work. Our goal is to extract the starting and ending time $(t_s, t_e)$, denoted as the answer span $a$, from the spoken passage $\mathbf{p}$ given the spoken question $\mathbf{q}$. 
\subsection{DUAL framework}
\begin{figure}[t]
    \centering
    \includegraphics[width=0.4\textwidth]{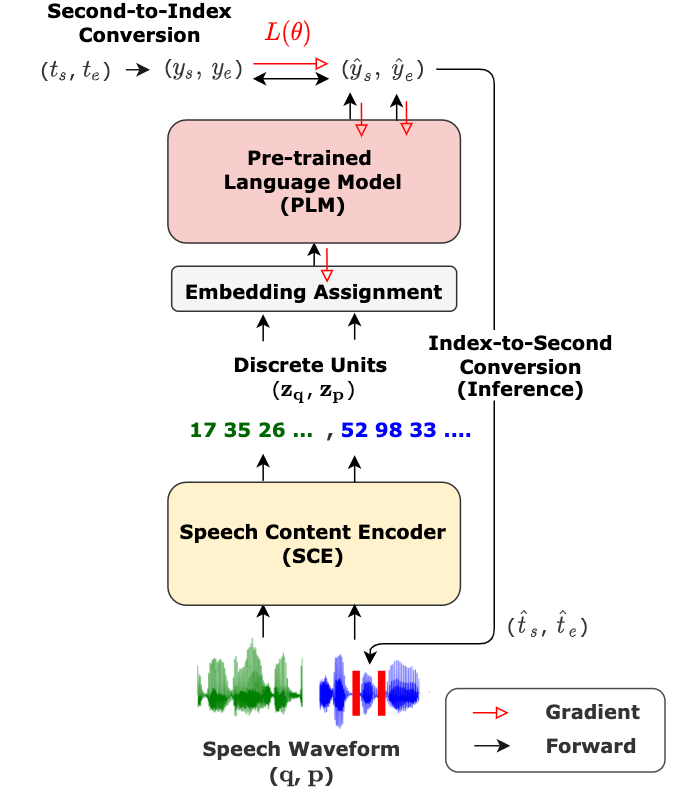}
    \caption{The overview of the DUAL framework.}
    \vspace{-0.5cm}
    \label{fig:DUAL}
\end{figure}

The DUAL framework consists of the Speech Content Encoder (SCE) and Pre-trained Language Model (PLM) as in Figure \ref{fig:DUAL} and introduced below.
\subsubsection{Speech Content Encoder (SCE)} 
The SCE transforms the question-passage audio waveform $(\mathbf{q, p})$ pair to sequences of discrete units $(\mathbf{z_q, z_p})$.\\
\textbf{Self-supervised Speech Representation}: A self-supervised speech pre-trained model can extract informative feature representations. We adopted the state-of-the-art self-supervised speech pre-trained model HuBERT \cite{hsu2021hubert} for feature extraction\footnote{We used the open-source S3PRL \cite{yang21c_interspeech} toolkit to extract the representations of the HuBERT-Large model.}. HuBERT was trained by masked prediction objectives similar to BERT \cite{devlin2018bert}. The prediction target was the K-means clustered index for speech signal processing features, e.g., Mel-frequency cepstral coefficients (MFCC) initially, and then learned latent representations after clustering in subsequent iterations. We utilized the HuBERT-Large model containing 24 transformer encoder layers pre-trained on LibriLight 60k hour dataset. HuBERT encoded the raw waveform into frame-level 1024 dimension features. Each frame was equivalent to 20 ms.\\
\textbf{Quantization}: The goal of quantization is to discretize the speech features so they can be fed into the pre-trained language model. K-means clustering was performed over the layer-wise representations of HuBERT-Large. We used LibriSpeech \cite{librispeech} 100-hour subset to train the K-means clustering model, and the number of clusters $K$ is 64, 128, and 512. After clustering, the discrete units are represented by the clustering indices. The repetitive discrete units are merged to shorten the sequence length and remove the duration information, forming the dense discrete unit sequence of the question and passage $(\mathbf{z_q}, \mathbf{z_p})$. We recorded the duration of duplication number of repetitions as $\mathbf{c_q}$ and $\mathbf{c_p}$ for $\mathbf{z_q}$ and $\mathbf{z_p}$, so we can recover the frame-level indices to convert the answer span back to time interval at the inference stage.
\looseness=-1
\subsubsection{Pre-trained Language Model (PLM)} 
The learning model is a BERT-like transformer encoder model. The input was the discrete unit sequences of the spoken questions and passages $(\mathbf{z_q}, \mathbf{z_p})$. Because SQA is a very challenging task to train from scratch, we leveraged the cross-disciplinary transferability of PLM \cite{jurafsky2020learning, kao2021bert, chiang2021transferability, lu2021pretrained, ri2022pretraining} to help the SQA downstream task. Specifically, we used the weights of text PLM for network initialization, and randomly assigned the text pre-trained input embeddings for discrete units. The different random embedding assignments did not significantly affect the final performance. The input of PLM was the concatenated discrete unit sequences of the question and passage pair $(\mathbf{z_q}, \mathbf{z_p})$, and the target was the start and end time $(y_s, y_e)$ after the repetitions were reproduced. 

Because the length of a discrete unit sequence is much longer than the corresponding text, and the duration of a spoken passage itself is long, the standard maximal length of PLM (typically 512) is not enough in our case. As a result, we leveraged Longformer \cite{beltagy2020longformer} to model the long $(\mathbf{z_q}, \mathbf{z_p})$, which is a BERT-like model for long documents, pre-trained on the unlabeled long text documents and optimized for training efficiency by sparse attention mechanism, such as local and global attention, to support up to $4096$ tokens.
\subsubsection{Training Objective}
\begin{table}[t]
\centering
\caption{Word Error Rates on different datasets for the two open-sourced ASR models used in the baseline (cascade).}
\vspace{-0.3cm}
\begin{tabular}{p{1.1cm}lp{1.7cm}lp{1.1cm}lp{1.1cm}}
\toprule
\textbf{ASR} & \textbf{LS test-clean} & \textbf{NMSQA dev} & \textbf{NMSQA test} \\
\midrule
SB & 3.1 & 15.6 & 61.7\\
W2v2 & 1.9 & 10.5 & 11.3\\
\bottomrule
\end{tabular}
\vspace{-0.4cm}
\label{tab:wer}
\end{table}
\vspace{-0.1cm}
The training objective is similar to the canonical QA fine-tuning in TQA. A randomly initialized linear layer is added on the top to predict the start and end time index. As the gradient flow shown in Figure \ref{fig:DUAL}, $\theta$ represents the trainable weights of the model, $\mathbf{c_p}=[\mathbf{c_p}_1, \mathbf{c_p}_2, ..., \mathbf{c_p}_n]$ is the repetition of every discrete units $\mathbf{z_p}_i$ in $\mathbf{z_p}=[\mathbf{z_p}_1, \mathbf{z_p}_2, ..., \mathbf{z_p}_n]$. $(t_s, t_e)$ is the ground truth start and end time in second, and $(y_s, y_e)$ is the converted the version on the index level. The overall training objective was to minimize the loss $L(\theta)$ as the sum of the negative log probabilities of the true start and end indices on all the examples,
\vspace{-0.1cm}
\begin{align*}
     - \sum \mathop{log}P(y_s|\mathbf{z_q}, \mathbf{z_p}; \theta) + \mathop{log}P(y_e|\mathbf{z_q}, \mathbf{z_p}; \theta).
\end{align*}

At the inference stage, we converted the predicted start and end indices $(\hat{y}_s, \hat{y}_e)$ to the frame level by $\mathbf{c_p}$ above, and finally transformed them to the time level $(\hat{t}_s, \hat{t}_e)$.
\section{Experiments}
\subsection{Corpus Description}
We developed and released here a new listening comprehension task named Natural Multi-speaker Spoken Question Answering (NMSQA). The train and dev set are the spoken version of the SQuAD v1.1 dataset, one of the largest QA datasets from Wikipedia paragraphs and human-written questions. We randomly split the SQuAD dev set into disjoint \textit{SQuAD-dev-1} and \textit{SQuAD-dev-2} for the NMSQA dev set and test set. The Amazon Polly Text-to-Speech service\footnote{https://aws.amazon.com/tw/polly/} was used for generating natural speech. We randomly assigned 12 TTS speakers and ensured that different speakers were used in producing each spoken document-question pair. Overall, there are 297.18 / 37.61 hours of audio for the train/dev set. Moreover, in order to have a realistic test set, 60 human speakers (30 male / 30 female) were requested to produce the \textit{SQuAD-dev-2} naturally. This test set included 2.67 hours of audio. The answer intervals were annotated by Montreal Force Aligner \cite{McAuliffe2017MontrealFA}. 
\subsection{Evaluation}
Since the output target here is the temporal span of the spoken answer, following the evaluation metrics previously proposed by \cite{li2018spoken, chuang2019speechbert}, we adopted the Frame-level F1 score (FF1) and Audio Overlapping Score (AOS) to evaluate the performance based on the predicted time intervals. Higher FF1 and AOS scores imply more overlap between the predicted and ground truth spans.

\begin{table}[t]
\centering
\caption{The performance of the proposed (DUAL) and baseline (cascade) approaches on the NMSQA dev and test sets. ``Longformer$^{\dagger}$" indicates the Longformer model fine-tuned on clean text SQuAD-v1.1, while the normal ``Longformer" was only pre-trained by unlabeled text data. The number after HuBERT (64, 128, 512) are numbers of clusters. ``synth" and ``human" represent the synthesized and human speech respectively.\looseness=-1}
\vspace{-0.3cm}
\begin{tabular}{lp{1.35cm}lp{0.55cm}lp{0.35cm}} 
\toprule
\multicolumn{1}{c}{\multirow{2}{*}{\textbf{Input}}} & \multirow{2}{*}{\textbf{Model}}  & \multicolumn{2}{c}{\textbf{dev (synth)}}                & \multicolumn{2}{c}{\textbf{test (human)}}          \\
\multicolumn{1}{c}{}                               &                                & \multicolumn{1}{c|}{FF1} & AOS                  & \multicolumn{1}{c|}{FF1} & AOS        \\
\midrule 
\multicolumn{6}{l}{\textbf{Baseline - Cascade (with ASR transcripts)}}                                                                                                                              \\
SB                                  & Longformer$^\dagger$   & 56.7                    & 49.7                & 17.3                    & 15.3         \\
W2v2                                      & Longformer$^\dagger$ &   65.7                       & 58.3 & 64.2 &  57.4\\
\midrule
\multicolumn{6}{l}{\textbf{\textbf{Proposed - DUAL (without ASR transcripts)}}}                                         \\
HuBERT-64                                          & Longformer                     & 47.8                    & 42.2               &      39.0                    & 33.0        \\
HuBERT-128                                         & Longformer                     & 54.2                  & 48.5               & \textbf{55.9}                    & \textbf{49.1}            \\
HuBERT-512                                         & Longformer                     &  \textbf{55.0}                        &  \textbf{49.6}                    &   17.3                       &       12.5        \\
\bottomrule
\vspace{-0.95cm}
\end{tabular}
\label{tab:overall_result}
\end{table}
\subsection{Baseline - Cascade (ASR plus TQA)}
The conventional SQA approach cascading an ASR model and a TQA model was taken as the baseline to be compared here. Two open-sourced pre-trained ASR models were used in the first stage, one from Speechbrain \cite{speechbrain}\footnote{https://huggingface.co/speechbrain/asr-crdnn-rnnlm-librispeech}, referred to as \textbf{SB}, the other the Wave2vec 2.0-large with self-training fine-tuning \cite{baevski2020wav2vec}\footnote{https://huggingface.co/facebook/wav2vec2-large-960h-lv60-self}, referred to as \textbf{W2v2}. The Word Error Rates of them on different speech datasets are listed in Table \ref{tab:wer}. Both SB and W2v2 utilized LibriSpeech \cite{librispeech} 960-hour dataset as the supervised training data; however, we see in Table \ref{tab:wer} W2v2 was much more robust than SB on the NMSQA test set, obviously because it leveraged 60k hrs of unlabeled data and the self-training procedure. 

The TQA model, or the second stage of the baseline, is a Longformer-based model fine-tuned on SQuAD v1.1, denoted as Longformer$^\dagger$ below. We used the online available model checkpoint\footnote{ https://huggingface.co/valhalla/longformer-base-4096-finetuned-squadv1} for TQA inference. The Longformer$^\dagger$ obtained 91.5 F1 score and 85.1 EM (Exact Match) score on the text SQuAD v1.1 dataset. For the evaluation metrics used here, we adopted force alignment \cite{McAuliffe2017MontrealFA} to obtain the time intervals of the spoken answers in seconds.

\subsection{Implementation Details of DUAL}
We use the official Longformer-base model\footnote{https://huggingface.co/allenai/longformer-base-4096} as the PLM. The learning rate is searched in [3e-5, 5e-5, 7e-5, 1e-4], and we select models with the best performance on the validation set. The learning rate warmup step is 500, growing up linearly to the peak value and then linearly decaying to 0. All the DUAL experiments use 4 Tesla V100s with an overall 128 batch size for up to 5000 training steps. If the length of discrete units $(\mathbf{z_q}, \mathbf{z_p})$ input exceeds 4096, we truncate the passage $\mathbf{z_p}$. 

\section{Results}
Encouraging results are reported here. Noting that the proposed (DUAL) approach achieves performance comparable to baselines (cascade), which require large ASR parallel data.
\subsection{``dev" set for synthesized speech} 
The experimental results are shown in Table \ref{tab:overall_result}. We first consider results for synthesized speech on the ``dev" set, whose style is very similar to the training set, in the first column.

The top section of Table \ref{tab:overall_result} is for the baseline (cascade) approach with ASR transcripts. We see ASR with W2v2 offered much better performance than that with SB (65.7 vs. 56.7 for FF1), obviously due to its lower recognition error in Table \ref{tab:wer}. On the other hand, for the proposed (DUAL) approach without ASR transcripts in the lower section of Table \ref{tab:overall_result}, we see DUAL achieved good FF1 scores (55.0, 54.2, 47.8 respectively for 512, 128, 64 clusters). Here, ``HuBERT-$K$'' denotes the input of DUAL is the clustering results from HuBERT-Large $22th$ layer representations with $K$ clusters. These numbers are competitive to those in the top section for the baseline, verifying DUAL can learn semantics in the audio signals and find the answers almost as good as the case where ASR transcripts were available, though transcripts were not available at all. The relatively weak performance for 64 clusters suggested that the too small codebook size may lose important fine-grained content information. The situation can be improved significantly using larger codebook sizes (128 or 512 clusters).
\subsection{``test" set for human speakers} 
The experimental results on a more realistic scenario of human speech (``test") are shown in the right column of Table \ref{tab:overall_result}. We observe that for the baseline (cascade) approach using SB for ASR on the top section, the performance dropped sharply due to the very high WER (61.7 in Table \ref{tab:overall_result}) on the human speech, while using the W2v2 for ASR model offered more robust results similar to the ``dev" set. The result indicated that the performance of the baseline (cascade) approach relied heavily on the accuracy and robustness of ASR. 
 
On the other hand, in the lower section of Table \ref{tab:overall_result}, we see the proposed (DUAL) approach could retain outstanding performance when K = 128, showing remarkable robustness of the approach for realistic human speech (55.9 of FF1 for K = 128). However, the performance dropped drastically for K = 512 (17.3 of FF1). The observation suggested that the cluster number played a crucial role in performance. We surmise that 128 clusters of quantization provide a smooth transformation from the machine-synthesized speech used in training to the human-speech test set. In contrast, 512 clusters may retain too many details that differentiate synthesized and human speech. This finding inspires more research to understand what makes textless SQA/NLP work or not.
\section{Analysis and Discussion}
\begin{figure}[t]
    \centering
    \includegraphics[width=0.36\textwidth]{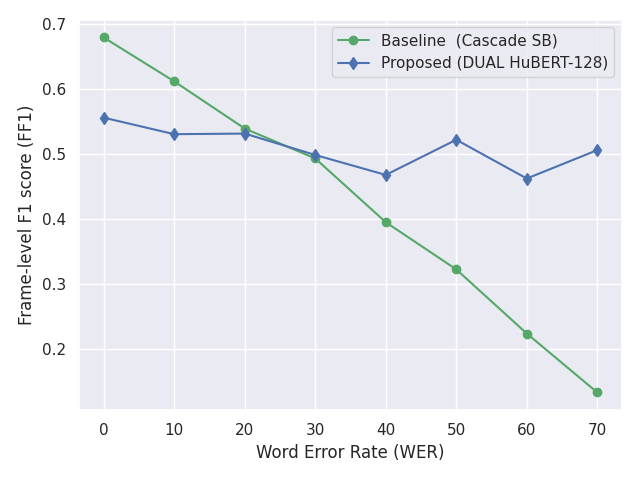}
    \vspace{-0.25cm}
    \caption{Frame-level F1 (FF1) scores for DUAL and cascade approach (SB), evaluated on the small groups of full NMSQA dev set at different levels of ASR (SB) WER.}
    \vspace{-0.5cm}
    \label{fig:wer_f1}
\end{figure}

\textbf{Ablation study for embedding assignment}: Table \ref{tab:emb} shows the ablation study regarding how the discrete units should be assigned to the pre-trained embeddings, as shown in the middle of Figure \ref{fig:DUAL}. In this table on the top two rows (``Most frequent" and ``Least frequent"), we randomly assigned the K (128) discrete units to the pre-trained embedding of the top-$K$ and the least-$K$ frequent vocabularies, where the vocabulary frequency was determined by Byte-Pair Encoding (BPE) on unlabeled text data. ``Random" refers to randomly selecting pre-trained input embedding regardless of the frequency. ``Re-init" denotes re-initializing the input embedding by a normal distribution. "Scratch" means the Longformer model was not pre-trained on the unlabeled text data. The results in Table \ref{tab:emb} indicate that randomly assigning the pre-trained input embeddings for discrete units did not result in very different performance, although the ``Most frequent" initialization offered the best results, which are those listed in Table \ref{tab:overall_result}. \\
\textbf{Performance for Poor ASR Accuracy}: We compared the performance of the baseline cascade approach (SB for ASR which gave poor accuracy) and the proposed DUAL with HuBERT-128 for different levels of WER. Specifically, we bucketize the NMSQA dev set into subsets based on the WER (from 0\% to 70\%) obtained with ASR (SB). In Figure \ref{fig:wer_f1}, we observe that for the baseline (cascade) approach, the FF1 score dropped significantly and continuously as the WER increased. This is the typical phenomenon of recognition error propagation. In contrast, the proposed (DUAL) attained very similar FF1 scores for different levels of WER, even when WER went up to 70 \%. Because there is no ASR in DUAL and no ASR transcripts were used, there was actually no correlation between WER and the FF1 score. The cascade approach outperformed DUAL when the WER was below 30\%,; but DUAL became much higher when WER exceeded 30\%. Since the content of SQuAD (and thus NMSQA) is based on Wikipedia, it includes many words that are name entities, abbreviations, and OOV, which led to recognition errors. Many of these words are part of the answers. DUAL handle such scenario much better than cascade approaches.
\begin{table}[t]
\centering
\caption{Ablation study on embedding assignment. All experiments used the HuBERT-128 setting. Performance was measured on the NMSQA dev set.}
\vspace{-0.3cm}
\begin{tabular}{llll}
\toprule
\textbf{Embedding Assignment} & \textbf{FF1} & \textbf{AOS}\\
\midrule
Most frequent & 54.2& 48.5\\
Least frequent & 46.9 & 41.7\\
Random & 51.7& 46.2\\
Re-init & 8.9 & 7.2 \\
\midrule
Scratch (baseline) & 6.1 & 4.9\\
\bottomrule
\end{tabular}
\vspace{-0.3cm}
\label{tab:emb}
\end{table}
\vspace{-0.1cm}
\section{Conclusion}
We propose DUAL, the first textless (i.e., ASR transcript-free) SQA framework in this work. This framework only utilizes unlabeled speech and text data for pre-training and fine-tuning by the spoken questions, passages, and answer time intervals. DUAL directly predicts the answer time span without text supervision or acoustic word boundaries. Furthermore, we release NMSQA, a new natural, multi-speaker SQA benchmark corpus, which contains human speakers for the test set and large-scaled synthesized data for the training and development sets. The experiments showed that DUAL yields competitive results with the conventional cascade approach using ASR transcripts and is robust to real-world scenarios on NMSQA.
\bibliographystyle{IEEEtran}

\bibliography{mybib}

\end{document}